\documentclass[11pt]{article}

\usepackage{pifont}
\usepackage[]{acl}
\usepackage{makecell}
\usepackage{times}
\usepackage{latexsym}
\usepackage{svg}
\usepackage{rotating}   
\usepackage{graphicx}
\usepackage{amssymb}
\DeclareRobustCommand{\cmark}{\textcolor{green}{\ding{51}}}  
\DeclareRobustCommand{\xmark}{\textcolor{red}{\ding{55}}} 
\usepackage[T1]{fontenc}
\usepackage{xcolor}   
\usepackage{amssymb} 
\usepackage[utf8]{inputenc}

\usepackage{microtype}
\usepackage[table]{xcolor}
\usepackage{amssymb}
\usepackage{inconsolata}
\usepackage{amsmath} 
\usepackage{graphicx}
\usepackage{xcolor}      
\usepackage{tcolorbox}   
\tcbuselibrary{skins, breakable}  

\newtcolorbox{promptcard}[1][]{
    colback=blue!8,           
    colframe=blue!40!gray,    
    coltitle=black,           
    fonttitle=\bfseries\large,
    sharp corners,            
    boxrule=0.8pt,            
    left=12pt, right=12pt, top=15pt, bottom=12pt,
    arc=0pt,                  
    auto outer arc,           
    breakable,                
    title={#1}                
}

\title{ParaRecover: A Process-Level Benchmark for Error Localization and Recovery in Parallel Tool-Use Agents}

\usepackage{xcolor}
\usepackage{soul}

\author{
Bowen Guan$^{1}$, Zhentao Yin$^{1}$, Yanming Shen$^{1}$\footnotemark[1] \\
$^{1}$School of Computer Science and Technology, Dalian University of Technology, China \\
\texttt{20201071138@mail.dlut.edu.cn, yinzhentao2003@mail.dlut.edu.cn}\\
\texttt{shen@dlut.edu.cn}
}

\usepackage{booktabs}   
\usepackage{tabularx}   
\usepackage{multirow}   
\usepackage{siunitx} 
\begin{document}
\maketitle
\begin{abstract}
Existing agent benchmarks mainly evaluate final task success or tool-call correctness, providing limited insight into whether agents can reliably diagnose and recover from intermediate execution failures. This limitation becomes particularly critical in multi-turn parallel tool-use scenarios, where errors may propagate across dependent branches and trigger cascading failures. We introduce ParaRecover, a process-level benchmark for evaluating error localization and recovery in multi-turn parallel tool-use agents. 
 Built upon a fine-grained taxonomy of 14 error types covering planning dependencies, tool selection, and argument matching, the benchmark comprises 10,626 instances spanning two difficulty levels.
To enable fine-grained, process-oriented evaluation, we further propose the SDE rubric, which measures structural integrity, diagnostic reasoning, and evolutionary strategy during agent execution. Experiments across more than ten mainstream LLMs reveal that even state-of-the-art models still struggle with multi-turn error propagation, implicit tool-use failures, and precise replanning. Moreover, we demonstrate that the SDE rubric provides effective supervision signals for improving agents’ reflective recovery capabilities. Our data and code are available at \url{https://github.com/gbw206/ParaRecover}.

\end{abstract}

\section{Introduction}
\begin{table*}
\centering
\small
\begin{tabular}{lcccccc}
\hline
 \textbf{Benchmark}  & \makecell{\textbf{Structured} \\ \textbf{Trajectory Eval.}} & \makecell{\textbf{Parallel} \\ \textbf{Calls}}
 & \makecell{\textbf{Error-State} \\ \textbf{Eval.}} & \makecell{\textbf{Reflective} \\\textbf{Localization}} & \makecell{\textbf{Replanning} \\ \textbf{Decision}} & \makecell{\textbf{Process-Level} \\ \textbf{Metrics}} \\
\hline
API-Bank  & \xmark & \xmark & \cmark & \xmark & \xmark & $\triangle$  \\
Tool-Bench  & \xmark & \xmark & $\triangle$ & \xmark & \xmark & $\triangle$  \\
BFCL  & \xmark & \cmark & \xmark & \xmark & \xmark & $\triangle$  \\
ToolSandbox  & \cmark & \cmark & \xmark & \xmark & \xmark & \cmark  \\
TRAJECT-Bench  & \cmark & \cmark & \xmark & \xmark & \xmark & \cmark \\
ParaRecover  & \cmark & \cmark & \cmark & \cmark & \cmark & \cmark  \\
\hline
\end{tabular}
\caption{Comparison between previous agent benchmarks and ParaRecover in evaluation dimensions. \protect\cmark{} denotes full support, \protect\xmark{} denotes no support, and $\triangle$ denotes indirect support.}
\label{comparison}
\end{table*}


Large language model agents are increasingly used to solve complex tasks through planning, tool use, and interaction with external environments~\citep{yao2022react, qin2024toolllm, huang2024understanding}. In real-world scenarios such as scientific research, code generation, and data analysis~\citep{ma2024sciagent, yang2024swe, hu2024infiagent}, tasks are often not simple sequential processes. Instead, they are usually multi-turn parallel processes, which involve multiple parallel subgoals, branching tool calls, and cross-step dependencies~\citep{huang2024understanding, wang2024survey}. Therefore, systematically evaluating agents’ planning, decision-making, and error-recovery abilities in multi-turn parallel tool-use scenarios has become an important problem.

Existing benchmarks for agent and tool use mainly focus on final task outcomes, such as answer correctness and overall task completion rate~\citep{mialon2024gaia, yao2024tau, liu2024agentbench}. However, successful task completion does not necessarily imply reliable agent behavior. An agent may still complete the task through redundant calls, repeated trial and error, or incorrect recovery strategies, leading to high cost in real systems. These end-to-end metrics are useful for measuring final performance, but they provide limited insight into whether agents can correctly localize failures, diagnose their causes, and recover from corrupted intermediate states.

As shown in Table~\ref{comparison}, although some recent studies consider process-level behaviors, they typically focus on tool-call success rate or provide only coarse-grained or indirect judgments of intermediate decisions or format~\citep{li2023api,qin2024toolllm, patil2025berkeley,  he2025traject, lu2025toolsandbox}. As a result, they may fail to capture the fine-grained behaviors required in multi-turn parallel execution, such as deciding concurrent subtasks, identifying cross-step dependencies, assessing the impact of failures on downstream branches, and replanning after errors.
Existing evaluation paradigms rarely characterize these recovery challenges in a systematic way, making it difficult to understand how and where agents fail. In practical applications, however, robust error diagnosis, reflection, and replanning capabilities are crucial for reliable long-horizon agent execution.



Motivated by these limitations,  we construct a fine-grained error taxonomy based on agent execution trajectories in real-world multi-turn parallel scenarios, covering 14 error types across planning dependencies, tool selection, and argument matching. Building upon this taxonomy, we develop ParaRecover, a benchmark designed for complex multi-turn parallel agent execution scenarios. ParaRecover covers two difficulty levels: LEVEL-1 only includes errors introduced in the most recent execution, while LEVEL-2 involves multi-turn propagation and various error types, comprehensively evaluating the agent's error localization and replanning capabilities. To further refine the evaluation, we introduce the \textbf{SDE} evaluation rubric, which measures agent behavior from three dimensions: \textbf{S}tructural Integrity,  \textbf{D}iagnostic Reasoning, and \textbf{E}volutionary Strategy. Our main contributions are as follows:
\begin{itemize}
\item We introduce a process-level benchmark for multi-turn parallel tool use, comprising 10,626 instances across two difficulty levels, explicitly evaluating agents’ intermediate decision-making, error localization, and corrective replanning.

\item We develop and validate a fine-grained error taxonomy for parallel agent execution. Based on real rollout trajectories from multiple models, we further characterize how these errors emerge and distribute in practice.

\item We propose the SDE Rubric, which evaluates agent from Structural Integrity, Diagnostic Reasoning, and Evolutionary Strategy, enabling fine-grained assessment of reflection, backtracking, and decision capability after errors.

\item We reveal a substantial gap between final task success and process reliability in current agents. Although state-of-the-art models often achieve high task completion rates, they still exhibit severe weaknesses in dependency reasoning, implicit failure localization, and efficient recovery under multi-turn error propagation.
\end{itemize}
\begin{figure*}[t]
  \includegraphics[width=\textwidth]{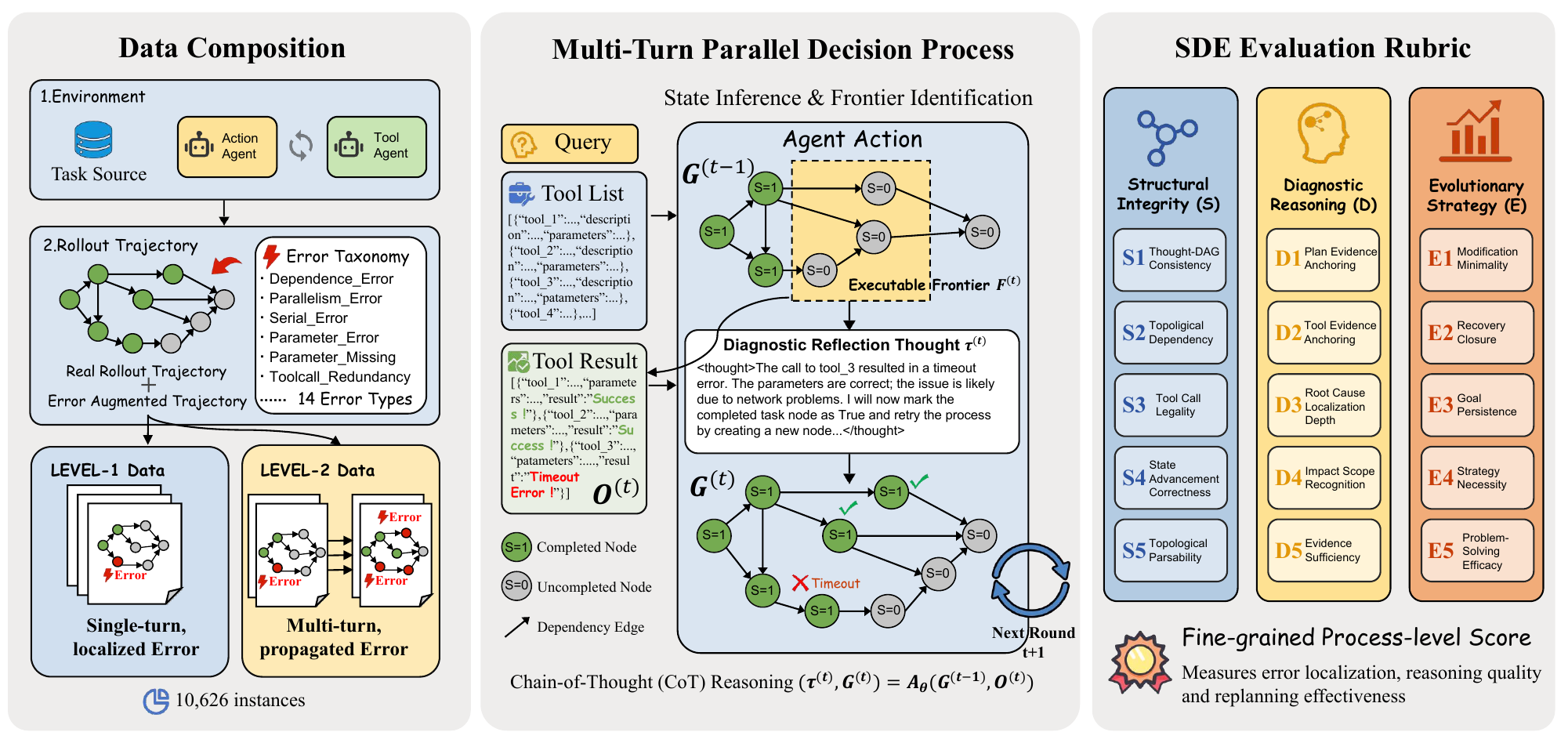}
  \caption{Overview of the ParaRecover Benchmark}
  \label{fig1}
\end{figure*}

\begin{table*}
\centering
\small
\begin{tabular}{l l c c c c c c c c c c}
\toprule
\multirow{2}{*}{\textbf{Family}} & \multirow{2}{*}{\textbf{Model Name}} & \multicolumn{5}{c}{\textbf{LEVEL-1}} & \multicolumn{5}{c}{\textbf{LEVEL-2}} \\
\cmidrule(lr){3-7} \cmidrule(lr){8-12}
& & \textbf{SI} & \textbf{DR} & \textbf{ES} & \textbf{Avg.} & \textbf{Pass@1} & \textbf{SI} & \textbf{DR} & \textbf{ES} & \textbf{Avg.} & \textbf{Pass@1}\\
\midrule
\multirow{4}{*}{OpenAI} & GPT-4o & 56.08 & 59.10 & 39.16 & 51.45 & 92.21 & 47.90 & 53.73 & 32.59 & 44.74 & 91.50\\
& GPT-4o mini & 49.91 & 46.20 & 29.83 & 41.98 & 89.32 & 44.85 & 40.41 & 25.45 & 36.90 & 88.84\\
& GPT-5 nano & 43.88 & 55.31 & 25.07 & 41.42& 89.50 & 40.53 & 51.61 & 22.56 & 38.23& 88.92 \\
& GPT-5.2 & 57.51 & \textbf{81.81} & 41.21 & 60.24& 92.61 & 55.24 & \textbf{80.75} & 41.61 & 59.20 & 91.84\\
\midrule
\multirow{2}{*}{Anthropic} & Claude Sonnet 4.6 & 69.12 & 74.36 & 50.27 & 64.58& 94.10 & 65.84 & 71.22 & 48.63 & 61.90 & 93.50\\
& Claude Opus 4.6 & 73.56 & 78.94 & 55.18 & \textbf{69.23} & \textbf{94.78}& 70.38 & 76.41 & 53.76 & \textbf{66.85} & \textbf{94.05}\\
\midrule
\multirow{2}{*}{Google} & Gemini 3.1 flash & 68.35 & 65.87 & 47.19 & 60.47& 93.82 & 63.50 & 61.13 & 45.04 & 56.56& 93.25 \\
& Gemini 3.1 pro & 66.90 & 61.72 & 48.23 & 58.95& 93.68 & 63.22 & 62.08 & 47.18 & 57.49& 93.22 \\
\midrule
\multirow{4}{*}{Qwen} 
& Qwen 3 8B & 47.61 & 57.08 & 25.72 & 43.47& 91.84 & 45.41 & 51.12 & 26.69 & 41.07 & 91.37\\
& Qwen 3 32B & 52.26 & 56.94 & 31.17 & 46.80& 92.08 & 50.42 & 54.55 & 31.94 & 45.64 & 91.86\\
& Qwen 3.5 9B & 56.58 & 72.78 & 38.35 & 55.90& 92.21 & 53.12 & 68.65 & 37.27 & 52.98 & 92.14\\
& Qwen 3.5 122B-A10B & 72.48 & 77.87 & 52.28 & 67.54& 94.50& 69.12 & 74.77 & 51.34 & 65.08& 93.04 \\
\midrule
\multirow{2}{*}{ZHIPU} & GLM 4.5 Air & 62.46 & 61.03 & 42.62 & 55.37 & 93.50& 58.43 & 59.33 & 41.67 & 53.14&92.72 \\
& GLM 4.6 & 66.20 & 55.91 & 47.80 & 56.64& 93.81 & 63.42 & 52.68 & 48.47 & 54.86 & 93.06 \\
\midrule
\multirow{3}{*}{DeepSeek} & DeepSeek-V4-flash & 73.74 & 75.80 & 54.57 & 68.04& 94.13 & 71.04 & 73.82 & 54.23 & 66.36 & 93.17\\
& DeepSeek-V4-flash* & 74.23 & 74.53 & 56.49 & 68.42& 94.67 & 71.50 & 72.90 & 55.30 & 66.57 & 93.86\\
& DeepSeek-V4-pro & \textbf{74.30} & 72.35 & \textbf{58.21} & 68.29& 94.13 & \textbf{71.47} & 71.10 & \textbf{57.03} & 66.50 & 93.01\\
\bottomrule
\end{tabular}
\caption{Performance of different models on LEVEL-1 and LEVEL-2 tasks}
\label{main}\end{table*}

\section{Problem Definition}

We consider real-world tasks, which contain a substantial number of subtasks that are conditionally independent and therefore can be executed in parallel. Although parallel execution can significantly improve efficiency, it also imposes substantially greater demands on agent planning and decision-making capabilities. To capture this more realistic and challenging setting, we formulate agent execution as a \textbf{multi-turn parallel decision process}.

To formalize parallelism, we represent the execution process as a DAG $G=(V,E)$, where each node $v_i\in V$ denotes a subtask or tool-use operation, and each edge $(v_i,v_j)\in E$ indicates that $v_j$ depends on the completion of $v_i$. Under this formulation, the prerequisite set of node $v_j$ can be written as
\begin{equation}
Dep(v_j) = \{ v_i\in V |(v_i,v_j)\in E \}.
\end{equation}
A node $v_j$ is executable only when all of its dependencies have been satisfied, i.e.,
\begin{equation}
\forall v_i \in Dep(v_j), \quad s(v_i) = 1,
\end{equation}
where $s(v_i) \in \{0,1\}$ represents the state of node $v_i$, indicating whether it has been completed. 
Consequently, tool invocations are executed layer-wise strictly along the topological structure of the DAG. Originating from the root, the graph naturally partitions into hierarchical layers: nodes directly dependent on the root constitute the first layer, nodes dependent on the first layer form the second, and so forth.

Since the topological depth reached in any given round is non-deterministic, the agent must dynamically infer the execution state. At each interaction step $t$, the agent receives the partially executed DAG $G^{(t-1)}$ from the previous round, along with the execution observations $O^{(t)}$ from the recently invoked tools. A critical capability required of the agent is to deduce the executable frontier for the current round. This frontier, denoted as $F^{(t)}$, comprises all pending nodes whose prerequisite dependencies have been strictly satisfied.

The core objective of the model at step $t$ is multifaceted: it must critically analyze the structural correctness of the preceding DAG, identify the target nodes within $F^{(t)}$, and evaluate the observations $O^{(t)}$ to diagnose which tool invocations succeeded or failed. To formalize this diagnostic reflection and decision-making process, we constrain the agent to follow a Chain-of-Thought (CoT)~\citep{wei2022chain} reasoning paradigm. Specifically, the agent $A_\theta$ is prompted to generate a textual reflection $\tau^{(t)}$ (thought) followed by an updated graph $G^{(t)}$, defined as the mapping:
\begin{equation}
(\tau^{(t)}, G^{(t)}) = A_\theta(G^{(t-1)}, O^{(t)}).
\end{equation}
The overall problem definition is shown in the middle part of Fig.~\ref{fig1}.

\subsection{ParaRecover}
As shown in Fig.~\ref{fig1}, ParaRecover is a process-level benchmark evaluating error localization and recovery in parallel tool-use agents via DAG.
\subsubsection{Task Source}

We source our initial tasks from BUTTON~\citep{chen2024facilitating}, which constructs complex trajectories by introducing dependencies among diverse atomic tasks. This compositional structure exhibits a natural correspondence with the edges and subtask nodes inherent in DAG planning. Regarding the tool execution environment, we follow~\citet{chen2024facilitating} and adopt the tool agent-based simulation paradigm. Since our primary objective is to evaluate the model's capacity for reflection and decision-making amidst various intermediate error states rather than assessing the stability of external APIs, this controlled simulated environment serves as an ideal execution backend.

\subsubsection{Error Taxonomy}
A core challenge is to determine what types of errors should be recovered. Previous work has identified several common failure modes~\citep{zhuang2023toolqa, chen2025acebench}, such as incorrect tool names, invalid arguments, and non-executable calls. However, these error types mainly capture local mistakes at the function-call level and do not fully characterize the failures that emerge in multi-turn parallel execution. In a DAG-based agent workflow, the model must additionally maintain valid dependency structures, identify which nodes can be executed in parallel and update the global plan after partial completion. 

Therefore, beyond inherited tool-level errors, we extract additional error categories from large-scale rollout traces, including parallelization errors, dependency errors, and so on. Finally, we construct a fine-grained error taxonomy consisting of three dimensions and 14 specific error types. Specific error types and their descriptions are provided in Appendix~\ref{Error Description}. We further report the empirical distribution of these errors across models of different parameter scales in Appendix~\ref{Error Type Distribution}, showing that they are not artificial annotations but naturally occurring failure patterns in realistic agent executions. We then use this unified error taxonomy as the basis for subsequent trajectory perturbation and error-aware data augmentation.

\begin{table*}
\centering
\small
\begin{tabular}{lccccccc}
\toprule
\multirow{2}{*}{\textbf{Error Type}} & \multicolumn{3}{c}{\textbf{LEVEL-1}} & \multicolumn{3}{c}{\textbf{LEVEL-2}} \\
\cmidrule(lr){2-4} \cmidrule(lr){5-7}
 & \textbf{DR} & \textbf{SI} & \textbf{ES} & \textbf{DR} & \textbf{SI} & \textbf{ES} \\
\midrule
\rowcolor{gray!12}
\multicolumn{7}{l}{\textbf{Structural-related Errors}} \\

Dependence\_Error      & 34.89 & 46.89 & 25.22 & $33.71_{\textcolor{red}{\tiny\downarrow 3.4\%}}$ & $43.80_{\textcolor{red}{\tiny\downarrow 6.6\%}}$  & $24.85_{\textcolor{red}{\tiny\downarrow 1.5\%}}$ \\

Dependence\_Null       & 39.34 & 48.15 & 22.20  & $34.61_{\textcolor{red}{\tiny\downarrow 12.0\%}}$ & $44.07_{\textcolor{red}{\tiny\downarrow 8.5\%}}$ & $23.05_{\textcolor{green}{\tiny\uparrow 3.8\%}}$ \\

Parallelism\_Error     & 34.19 & 54.26 & 29.79 & $32.53_{\textcolor{red}{\tiny\downarrow 4.9\%}}$  & $47.48_{\textcolor{red}{\tiny\downarrow 12.5\%}}$ & $26.83_{\textcolor{red}{\tiny\downarrow 9.9\%}}$ \\

Serial\_Error          & 41.26 & 49.78 & 26.24 & $33.99_{\textcolor{red}{\tiny\downarrow 17.6\%}}$ & $48.09_{\textcolor{red}{\tiny\downarrow 3.4\%}}$  & $26.26_{\textcolor{green}{\tiny\uparrow 0.01\%}}$ \\

\rowcolor{gray!12}
\multicolumn{7}{l}{\textbf{Parameter-related Errors}} \\

Parameter\_Error       & 57.35 & 51.95 & 32.80  & $48.82_{\textcolor{red}{\tiny\downarrow 14.9\%}}$ & $44.88_{\textcolor{red}{\tiny\downarrow 13.6\%}}$ & $27.05_{\textcolor{red}{\tiny\downarrow 17.5\%}}$ \\

Parameter\_Invalid     & 49.63 & 49.21 & 29.95 & $45.67_{\textcolor{red}{\tiny\downarrow 8.0\%}}$  & $44.82_{\textcolor{red}{\tiny\downarrow 8.9\%}}$  & $28.51_{\textcolor{red}{\tiny\downarrow 4.8\%}}$ \\

Parameter\_Missing     & 57.93 & 49.63 & 31.90  & $48.53_{\textcolor{red}{\tiny\downarrow 16.2\%}}$ & $45.86_{\textcolor{red}{\tiny\downarrow 7.6\%}}$  & $29.81_{\textcolor{red}{\tiny\downarrow 6.6\%}}$ \\

Type\_Error            & 55.19 & 49.04 & 31.56 & $46.68_{\textcolor{red}{\tiny\downarrow 15.4\%}}$ & $47.71_{\textcolor{red}{\tiny\downarrow 2.7\%}}$  & $31.18_{\textcolor{red}{\tiny\downarrow 1.2\%}}$ \\

\rowcolor{gray!12}
\multicolumn{7}{l}{\textbf{Tool-related Errors}} \\

Null\_Error            & 46.12 & 44.48 & 23.88 & $37.74_{\textcolor{red}{\tiny\downarrow 18.2\%}}$ & $37.20_{\textcolor{red}{\tiny\downarrow 16.4\%}}$  & $17.14_{\textcolor{red}{\tiny\downarrow 28.2\%}}$ \\

Timeout\_Error         & 44.17 & 41.98 & 21.57 & $37.28_{\textcolor{red}{\tiny\downarrow 15.6\%}}$ & $37.27_{\textcolor{red}{\tiny\downarrow 11.2\%}}$ & $16.39_{\textcolor{red}{\tiny\downarrow 24.0\%}}$ \\

Tool\_name\_Error      & 41.91 & 35.03 & 16.80  & $35.61_{\textcolor{red}{\tiny\downarrow 15.0\%}}$ & $34.99_{\textcolor{red}{\tiny\downarrow 0.1\%}}$  & $17.72_{\textcolor{green}{\tiny\uparrow 5.5\%}}$ \\

Tool\_select\_Error    & 42.07 & 52.66 & 28.98 & $40.38_{\textcolor{red}{\tiny\downarrow 4.0\%}}$  & $44.64_{\textcolor{red}{\tiny\downarrow 15.2\%}}$ & $22.64_{\textcolor{red}{\tiny\downarrow 21.9\%}}$ \\

Toolcall\_Missing      & 35.81 & 54.35 & 26.65 & $37.11_{\textcolor{green}{\tiny\uparrow 3.6\%}}$  & $32.36_{\textcolor{red}{\tiny\downarrow 40.5\%}}$ & $16.91_{\textcolor{red}{\tiny\downarrow 36.6\%}}$ \\

Toolcall\_Redundancy   & 31.85 & 52.01 & 27.06 & $27.99_{\textcolor{red}{\tiny\downarrow 12.1\%}}$ & $29.79_{\textcolor{red}{\tiny\downarrow 42.7\%}}$ & $18.36_{\textcolor{red}{\tiny\downarrow 32.2\%}}$ \\

\bottomrule
\end{tabular}
\caption{Performance of GPT-4o-mini across different error types}
\label{error:4omini}
\end{table*}

\subsubsection{Data Composition}
Our benchmark consists of two complementary sources: manually curated real rollout trajectories with errors and augmented error trajectories constructed by injecting the 14 predefined error types into correct trajectories. The former ensures that the benchmark reflects naturally occurring failure patterns in realistic agent rollouts, while the latter enables controlled expansion of error diversity and difficulty.

Based on the propagation range of errors, we organize the benchmark into two difficulty levels, namely LEVEL-1 and LEVEL-2. LEVEL-1 contains both manually selected and augmented single-round error trajectories, where the error is restricted to the immediately preceding round, either in the planning result or in the tool return. In this setting, the error has not yet propagated across multiple rounds, and the agent is mainly required to identify and correct a recent error. 

In contrast, LEVEL-2 is designed to capture more difficult failure cases. In practice, early errors are often not detected immediately. Instead, they continue to affect later execution. Moreover, an agent may introduce new errors in different rounds, and the accumulation of errors can jointly cause the final failure. LEVEL-2 is intended to model such conditions: multiple co-existing errors and cross-round error propagation. Specifically, from real rollout trajectories, we extract instances where errors are not immediately resolved but propagate to subsequent rounds, or where new errors are introduced later. For augmented data, we randomly inject various error types into the downstream steps of already flawed trajectories. So this level evaluates not only whether the agent can detect an error, but also recover under long-horizon corrupted contexts. Detailed data distributions are provided in Appendix~\ref{Data Distribution}.

\subsubsection{The SDE Rubric}
To overcome the limitations of traditional binary metrics in evaluating complex agent trajectories, we propose a fine-grained, multi-dimensional evaluation framework: the \textbf{SDE} (\textbf{S}tructural, \textbf{D}iagnostic, and \textbf{E}volutionary) Rubric. These three sub-dimensions correspond respectively to multi-turn parallel execution, error localization, and replanning capabilities.

\textbf{Structural Integrity (S)}
This dimension primarily evaluates the formal correctness and executability of the generated DAG, and is divided into 5 sub-dimensions ($S1-S5$).

\textbf{S1}: Thought-DAG Consistency (1/0.5/0): Evaluates whether the reasoning expressed in thought $\tau$ is perfectly mapped onto the generated DAG $\theta(G)$.

\textbf{S2}: Topological Dependency (1/0.5/0): Assesses the correctness of node dependencies, ensuring that parallel nodes have no interdependencies and sequential nodes have valid causal links.

\textbf{S3}: Tool Call Legality (1/0.5/0): Verifies the validity of the invoked tools and their parameters.

\textbf{S4}: State Advancement Correctness (1/0.5/0): Checks whether the status attributes of all nodes are accurately updated based on the prior execution feedback.

\textbf{S5}:Topological Parsability (1/0): Determines whether the generated DAG structure can be successfully parsed by the execution engine.

\begin{table*}
\centering
\small
\begin{tabular}{lcccccc}
\toprule
\multirow{2}{*}{\textbf{Error\_Type}} & \multicolumn{3}{c}{\textbf{LEVEL-1}} & \multicolumn{3}{c}{\textbf{LEVEL-2}} \\
\cmidrule(lr){2-4} \cmidrule(lr){5-7}
 & \textbf{DR} & \textbf{SI} & \textbf{ES} & \textbf{DR} & \textbf{SI} & \textbf{ES} \\
\midrule

\rowcolor{gray!12}
\multicolumn{7}{l}{\textbf{Structural-related Errors}} \\

Dependence\_Error      & 72.56 & 79.22 & 53.37 & $71.71_{\textcolor{red}{\tiny\downarrow 1.2\%}}$ & $73.45_{\textcolor{red}{\tiny\downarrow 7.3\%}}$ & $52.23_{\textcolor{red}{\tiny\downarrow 2.1\%}}$ \\

Dependence\_Null       & 74.02 & 76.09 & 50.87 & $72.75_{\textcolor{red}{\tiny\downarrow 1.7\%}}$ & $70.88_{\textcolor{red}{\tiny\downarrow 6.8\%}}$ & $51.60_{\textcolor{green}{\tiny\uparrow 1.4\%}}$ \\

Parallelism\_Error     & 71.12 & 81.06 & 56.12 & $69.46_{\textcolor{red}{\tiny\downarrow 2.3\%}}$ & $73.14_{\textcolor{red}{\tiny\downarrow 9.8\%}}$ & $55.44_{\textcolor{red}{\tiny\downarrow 1.2\%}}$ \\

Serial\_Error          & 73.55 & 78.42 & 55.86 & $71.77_{\textcolor{red}{\tiny\downarrow 2.4\%}}$ & $70.74_{\textcolor{red}{\tiny\downarrow 9.8\%}}$ & $54.09_{\textcolor{red}{\tiny\downarrow 3.2\%}}$ \\

\rowcolor{gray!12}
\multicolumn{7}{l}{\textbf{Parameter-related Errors}} \\

Parameter\_Error       & 84.32 & 67.11 & 54.39 & $80.47_{\textcolor{red}{\tiny\downarrow 4.6\%}}$ & $70.08_{\textcolor{green}{\tiny\uparrow 4.4\%}}$ & $55.50_{\textcolor{green}{\tiny\uparrow 2.0\%}}$ \\

Parameter\_Invalid     & 79.74 & 72.11 & 54.21 & $79.03_{\textcolor{red}{\tiny\downarrow 0.9\%}}$ & $72.26_{\textcolor{green}{\tiny\uparrow 0.2\%}}$ & $56.82_{\textcolor{green}{\tiny\uparrow 4.8\%}}$ \\

Parameter\_Missing     & 81.80 & 63.92 & 52.59 & $79.84_{\textcolor{red}{\tiny\downarrow 2.4\%}}$ & $68.22_{\textcolor{green}{\tiny\uparrow 6.7\%}}$ & $54.36_{\textcolor{green}{\tiny\uparrow 3.4\%}}$ \\

Type\_Error            & 83.10 & 68.40 & 55.24 & $79.68_{\textcolor{red}{\tiny\downarrow 4.1\%}}$ & $67.15_{\textcolor{red}{\tiny\downarrow 1.8\%}}$ & $56.04_{\textcolor{green}{\tiny\uparrow 1.4\%}}$ \\

\rowcolor{gray!12}
\multicolumn{7}{l}{\textbf{Tool-related Errors}} \\

Null\_Error            & 73.13 & 72.24 & 54.81 & $70.77_{\textcolor{red}{\tiny\downarrow 3.2\%}}$ & $70.71_{\textcolor{red}{\tiny\downarrow 2.1\%}}$ & $51.21_{\textcolor{red}{\tiny\downarrow 6.6\%}}$ \\

Timeout\_Error         & 71.51 & 68.07 & 53.21 & $70.74_{\textcolor{red}{\tiny\downarrow 1.1\%}}$ & $71.57_{\textcolor{green}{\tiny\uparrow 5.1\%}}$ & $52.50_{\textcolor{red}{\tiny\downarrow 1.3\%}}$ \\

Tool\_name\_Error      & 76.89 & 63.77 & 52.75 & $74.71_{\textcolor{red}{\tiny\downarrow 2.8\%}}$ & $64.35_{\textcolor{green}{\tiny\uparrow 0.9\%}}$ & $52.84_{\textcolor{green}{\tiny\uparrow 0.2\%}}$ \\

Tool\_select\_Error    & 75.85 & 81.70 & 56.01 & $73.80_{\textcolor{red}{\tiny\downarrow 2.7\%}}$ & $72.21_{\textcolor{red}{\tiny\downarrow 11.6\%}}$ & $53.74_{\textcolor{red}{\tiny\downarrow 4.1\%}}$ \\

Toolcall\_Missing      & 74.03 & 81.34 & 58.98 & $72.39_{\textcolor{red}{\tiny\downarrow 2.2\%}}$ & $75.86_{\textcolor{red}{\tiny\downarrow 6.7\%}}$ & $53.98_{\textcolor{red}{\tiny\downarrow 8.5\%}}$ \\

Toolcall\_Redundancy   & 69.31 & 83.44 & 55.53 & $66.13_{\textcolor{red}{\tiny\downarrow 4.6\%}}$ & $77.41_{\textcolor{red}{\tiny\downarrow 7.2\%}}$ & $52.84_{\textcolor{red}{\tiny\downarrow 4.9\%}}$ \\

\bottomrule
\end{tabular}
\caption{Performance of DeepSeek-V4-flash across different error types}
\label{error:dsv4}
\end{table*}

\textbf{Diagnostic Reasoning (D)}
This dimension assesses the depth and accuracy of the agent's failure analysis, and is divided into 5 sub-dimensions ($D1-D5$).

\textbf{D1}: Plan Evidence Anchoring (1/0.5/0): Evaluates whether the thought $\tau$ demonstrates a correct understanding and accurate referencing of the previous round's planning graph.

\textbf{D2}: Tool Evidence Anchoring (1/0.5/0): Assesses whether the thought $\tau$ accurately interprets the methods and execution results of previous tool calls.

\textbf{D3}: Root Cause Localization Depth (1/0.5/0): Determines if the reasoning identifies the fundamental root cause of the failure rather than merely describing surface-level symptoms.

\textbf{D4}: Impact Scope Recognition (1/0.5/0): Checks if the agent successfully identifies downstream nodes and dependency chains affected by a specific error.

\textbf{D5}: Evidence Sufficiency (1/0.5/0): Evaluates whether the reasoning provides adequate justification while avoiding over-inference.

\textbf{Evolutionary Strategy (E)}
This dimension measures the efficiency and precision of the agent's replanning, and is divided into 5 sub-dimensions ($E1-E5$).

\textbf{E1}: Modification Minimality (1/0.5/0): Assesses whether the updated plan only alters necessary components without introducing irrelevant revisions.

\textbf{E2}: Recovery Closure (1/0.5/0): Evaluates whether the agent not only fixed the explicit point of failure, but also the previous or subsequent faulty workflows associated with it.

\textbf{E3}: Goal Persistence (1/0.5/0): Verifies that the new plan remains strictly aligned with the original task objective without goal drift.

\textbf{E4}: Strategy Necessity (1/0.5/0): Determines whether the new step has a clear and reasonable necessity.

\textbf{E5}: Problem-Solving Efficacy: Quantifies the practical utility of the updated plan when executed in a real-world environment. It measures whether the task can be completed within finite steps and evaluates the margin of difference from the optimal trajectory. For each query $q$, we define the following variables: $G_q$, which is the optimal number of rounds required by the ground truth, $T_q$, which is the actual number of rounds required to complete the task after substituting the current plan into the subsequent workflow, and $S_q \in \{0, 1\}$, which is the final success status, where 1 indicates success and 0 indicates failure. The resulting score $Score(E5)$ for the query is defined as:
\begin{equation}
Score(E5) = S_q \cdot \min \left( 1, \frac{G_q}{T_q} \right).
\end{equation}

\section{Experiment}
We comprehensively evaluate agents’ reflection and decision-making capabilities under erroneous states on our benchmark. Specifically, we focus on the following research questions:

\textbf{RQ1}: How do mainstream LLMs perform on our benchmark?

\textbf{RQ2}: How does agent performance vary across different types of errors?

\textbf{RQ3}: Is it reliable to evaluate abstract rubrics using a LLM-as-a-Judge paradigm?

\textbf{RQ4}: Can fine-grained evaluation metrics provide actionable guidance for improving model capabilities?

\subsection{RQ1: How do mainstream large language models perform on our benchmark?}

We benchmark six major model families: (1) the OpenAI models, including GPT-4o~\citep{hurst2024gpt}, GPT-4o-mini~\citep{hurst2024gpt}, GPT-5-nano~\citep{singh2025openai}, and GPT-5.2~\citep{openai2025gpt52}; (2) the Anthropic models, including Claude Sonnet 4.6~\citep{claude46sonnet} and Claude Opus 4.6~\citep{claudeopus46}; (3) the Google Gemini models, including Gemini-3.1-Flash~\citep{gemini31flash} and Gemini-3.1-Pro~\citep{gemini31}; (4) the Qwen models, including Qwen3-8B, Qwen3-32B~\citep{yang2025qwen3}, Qwen3.5-9B, and Qwen3.5-122B-A10B~\citep{qwen2026qwen35}; (5) the Zhipu GLM models, including GLM-4.5-Air~\citep{zeng2025glm} and GLM-4.6~\citep{glm46}; and (6) the DeepSeek models, including DeepSeek-V4-Flash, DeepSeek-V4-Flash*(thinking mode), and DeepSeek-V4-Pro~\citep{deepseekv4}.

\textbf{Settings.} For all models, we adopt a consistent CoT prompt (see Appendix~\ref{prompt}) to guide model reasoning and decision-making. Evaluations are conducted under the SDE rubric, with results reported in Table~\ref{main}. In the table, SI, DR, and ES denote the average scores for Structural Integrity, Diagnostic Reasoning, and Evolutionary Strategy, respectively. The five sub-dimensions within each category are equally weighted.  All reported scores are normalized to the 0\~{}100 scale for ease of interpretation. Meanwhile, we report the pass@1 metric, where the updated plan will be adopted in subsequent rounds. A score of 1 is assigned upon successful task completion within specified steps (see Appendix~\ref{times}), and 0 otherwise.

The results show that mainstream LLMs achieve average scores below 70 on our benchmark. Moreover, models with larger parameter sizes substantially outperform smaller ones, indicating that our evaluation effectively captures differences in model capability. The generally low scores also suggest that state-of-the-art models still struggle with attribution and decision-making in multi-turn parallel tasks involving errors. Models consistently achieve higher scores across all dimensions on LEVEL-1 than on LEVEL-2, demonstrating that multi-turn errors pose a greater challenge than single-turn errors. Among all models, Claude Opus 4.6 achieves the highest average scores on both levels, obtaining 69.23 and 66.85, respectively.
Further analysis reveals that low scores are most frequently observed in the Evolutionary Strategy dimension, which measures a model’s ability to replan effectively in response to errors. While models tend to be relatively proficient at localizing errors, they show limited capacity for generating precise and efficient plan updates.

\begin{table}
\centering
\scriptsize
\begin{tabular}{cccccc}
\toprule
\textbf{Dimension} & \textbf{EMA} & \textbf{MAE} & \textbf{Human Mean} & \textbf{LLM Mean}\\
\midrule
\rowcolor{gray!12}
\multicolumn{5}{l}{Diagnostic Reasoning} \\
D1 & 0.8225 & 0.1062 & 0.6363 & 0.6700 \\
D2 & 0.8125 & 0.1100 & 0.5550 & 0.5725 \\
D3 & 0.8275 & 0.1100 & 0.5725 & 0.5775 \\
D4 & 0.8250 & 0.1162 & 0.5300 & 0.5688 \\
D5 & 0.8100 & 0.0950 & 0.3250 & 0.3950 \\
\midrule
\rowcolor{gray!12}
\multicolumn{5}{l}{Evolutionary Strategy} \\
E1 & 0.8125 & 0.1150 & 0.4475 & 0.4625 \\
E2 & 0.8025 & 0.0988 & 0.2350 & 0.2463 \\
E3 & 0.7950 & 0.1088 & 0.4238 & 0.3925 \\
E4 & 0.7925 & 0.1338 & 0.4963 & 0.5100 \\
\midrule
\rowcolor{gray!12}
\multicolumn{5}{l}{Structural Integrity} \\
S1 & 0.8275 & 0.1075 & 0.5125 & 0.5150 \\
S2 & 0.7875 & 0.1425 & 0.4750 & 0.4725 \\
S3 & 0.7875 & 0.1562 & 0.4938 & 0.4775 \\
S4 & 0.7850 & 0.1700 & 0.4538 & 0.3963 \\
\midrule
Overall & 0.8067 & 0.1208 & 0.4736 & 0.4813 \\
\bottomrule
\end{tabular}
\caption{Consistency between human annotations and LLM-as-a-Judge}
\label{tab:performance}
\end{table}

We can also observe that Pass@1 of all models is above 89\%, indicating that the plans generated by agents can mostly complete the tasks in subsequent rounds. However, all models score poorly on ES (where E5 measures the deviation from optimal execution rounds), revealing severe inefficiencies caused by redundant calls and execution errors. This highlights that final-outcome metrics alone fail to capture intermediate planning quality, whereas our evaluation system successfully identifies these process-level deficiencies. For further details, please refer to Appendix~\ref{Analysis of the Discrepancy Between Pass@1 and ES Scores}

\subsection{RQ2: How does agent performance vary across different types of errors?}
To investigate which types are difficult to localize or recover, we evaluate two representative models with contrasting parameter scales—GPT-4o-mini (small) and DeepSeek-V4-flash (large)—on tasks labeled with different error types. Experimental results are presented in Table~\ref{error:4omini} and Table~\ref{error:dsv4}, reporting overall dataset metrics for LEVEL-1 (single-turn) and LEVEL-2 (multi-turn). Red/green arrows indicate the relative change of metrics on LEVEL-2 compared to LEVEL-1.

Overall, multi-turn errors are more challenging than single-turn errors for both small and large models. Notably, DeepSeek-V4-flash shows relatively small changes from LEVEL-1 to LEVEL-2 across most metrics, with some dimensions even showing slight improvements, whereas GPT-4o-mini suffers substantial drops on most dimensions. This suggests that models with stronger reasoning capabilities can better handle multi-turn error scenarios and remain robust.

For both models, Toolcall\_Redundancy has the lowest scores in the DR dimension, indicating that redundant call is the hardest to identify. Unlike other errors, it typically does not cause explicit execution failures but manifest as “plausible yet ineffective” implicit errors, making them difficult to detect—a common challenge for agents.

\begin{table}
\centering
\scriptsize
\begin{tabular}{lcccc}
\toprule
\textbf{Model}
& \multicolumn{2}{c}{\textbf{LEVEL-1}}
& \multicolumn{2}{c}{\textbf{LEVEL-2}} \\
\cmidrule(lr){2-3}
\cmidrule(lr){4-5}
& DeepSeek & GPT-4o
& DeepSeek & GPT-4o \\
\midrule
GPT-4o
& 51.45 & 52.31
& 44.74 & 45.41 \\

Claude Opus 4.6
& 69.23 & 69.20
& 66.85 & 67.24 \\

Qwen3-8B
& 43.47 & 43.76
& 41.07 & 41.34 \\

DeepSeek-V4-flash
& 68.04 & 68.47
& 66.36 & 67.67 \\

DeepSeek-V4-pro
& 68.29 & 68.87
& 66.50 & 67.36 \\
\bottomrule
\end{tabular}
\caption{Cross-judge comparison of average SDE scores using
DeepSeek-V4-flash thinking and GPT-4o as judges.}
\label{tab:cross_judge}
\end{table}

Different error types impose distinct challenges. Parameter-related errors yield relatively high scores for two models. However, when difficulty escalates to LEVEL-2, GPT-4o-mini experiences a broad decline, while DeepSeek-V4-flash shows few declines and even improvements. This indicates that large model has become insensitive to this error type, effectively localizing and correcting them in both simple and complex calls.

Structure-related errors and tool-related errors pose significant challenges for both models. Models score relatively higher on parameter-related explicit errors. In real-world scenarios, explicit failures are often easy to resolve, whereas errors that accumulate implicitly—such as redundant calls, state drift, and unreasonable planning—remain difficult to handle.

\begin{table*}
  \centering
  \small
  \begin{tabular}{lcccccccc}
    \toprule
    \multirow{2}{*}{\textbf{Model}} & \multicolumn{4}{c}{\textbf{LEVEL-1}} & \multicolumn{4}{c}{\textbf{LEVEL-2}} \\
    \cmidrule(lr){2-5} \cmidrule(lr){6-9}
                           & \textbf{SI}  & \textbf{DR}  & \textbf{ES}  & \textbf{Avg.} & \textbf{SI}  & \textbf{DR}  & \textbf{ES}  & \textbf{Avg.} \\
    \midrule
    Base(Qwen3-8B)         & 47.35 &  57.75   &  26.48   & 43.86    &  46.33    &  51.06   &  27.72   &  41.70\\
    Few-shot           &    48.75  & 57.82 &  26.83 &  44.47  &    47.62 & 51.20 & 27.91 &     42.24      \\
    SFT             &    49.35  & 58.42 &  27.05 &  44.94  &    48.53 & 51.73 & 27.85 &     42.70      \\
    
    ParaRecover-DPO                 & \textbf{53.42}  &  \textbf{62.71}  &   \textbf{31.16} &  $\textbf{49.10}_{\textcolor{green}{\tiny\uparrow 5.24}}$    & \textbf{53.02}    &   \textbf{55.27}  &    \textbf{30.95} &   $\textbf{46.41}_{\textcolor{green}{\tiny\uparrow 4.71}}$   \\
    \bottomrule
  \end{tabular}
  \caption{Model Performance Comparison}
  \label{RL}
\end{table*}

\subsection{RQ3: Is it reliable to evaluate abstract rubrics using a LLM-as-a-Judge paradigm?}
Many dimensions in our evaluation rubric rely on abstract criteria, making manual annotation impractical. We therefore adopt LLM‑as‑a‑Judge, leveraging the reasoning capability of large language models to automatically score these abstract dimensions. Prior work shows that fine‑grained, low‑cardinality, and easily discernible rules substantially improve accuracy and consistency~\citep{hashemi2024llm, zhuge2024agent}, and our three‑level (1/0.5/0) scoring per dimension aligns well with this principle.

After evaluating multiple LLMs in preliminary experiments, we selected DeepSeek‑V4‑flash thinking, which achieved the highest initial agreement with human annotations. For each of the 13 LLM‑judged dimensions, we randomly sampled 400 instances, conducted multiple rounds of human annotation, and iteratively refined the evaluation prompt based on this standard. Four metrics are used to measure human‑model agreement: EMA, MAE, Human Mean, and Model Mean.

EMA (Exact Match Accuracy) is defined as:
\begin{equation}
\text{EMA} = \frac{1}{N} \sum_{i=1}^{N} \mathbf{1}_{\hat{y}_i = y_i},
\end{equation}
where $N$ is the total number of samples, $\hat{y}_i$ and $y_i$ are the model-predicted and human‑annotated scores, and $\mathbf{1}$ is the indicator function. 
MAE (Mean Absolute Error) is defined as:
\begin{equation}
\text{MAE} = \frac{1}{N} \sum_{i=1}^{N} \left| \hat{y}_i - y_i \right|.
\end{equation}

As reported in Table~\ref{tab:performance}, the average EMA across all dimensions is 0.8067, and the average MAE is 0.1208, demonstrating that the LLM evaluations achieve a trustworthy level of agreement. The evaluation prompt is provided in the Appendix~\ref{prompt}. 

\begin{table*}[t]
\centering
\resizebox{\textwidth}{!}{
\begin{tabular}{lcccccccc}
\toprule
\multirow{2}{*}{\textbf{Model}}
& \multicolumn{4}{c}{\textbf{LEVEL-1}}
& \multicolumn{4}{c}{\textbf{LEVEL-2}} \\
\cmidrule(lr){2-5}
\cmidrule(lr){6-9}
& \textbf{Tool Use Eff.$\uparrow$}
& \textbf{E5$\uparrow$}
& \textbf{Invalid DAG Rate$\downarrow$}
& \textbf{Pass@1$\uparrow$}
& \textbf{Tool Use Eff.$\uparrow$}
& \textbf{E5$\uparrow$}
& \textbf{Invalid DAG Rate$\downarrow$}
& \textbf{Pass@1$\uparrow$} \\
\midrule
Base (Qwen3-8B)
& 20.43 & 36.50 & 10.25 & 91.84
& 17.07 & 30.21 & 13.22 & 91.37 \\

SFT
& 21.29 & 37.12 & 6.71 & 92.42
& 18.42 & 31.64 & 9.41 & 92.02 \\

ParaRecover-DPO
& \textbf{26.20} & \textbf{40.85} & \textbf{5.90} & \textbf{94.30}
& \textbf{21.31} & \textbf{33.10} & \textbf{9.08} & \textbf{93.97} \\
\bottomrule
\end{tabular}
}
\caption{Judge-free evaluation of Base, SFT, and ParaRecover-DPO models on LEVEL-1 and LEVEL-2. Higher is better for Tool Use Efficiency, E5, and Pass@1, while lower is better for Invalid DAG Rate.}
\label{tab:judge_free}
\end{table*}

To further examine the sensitivity to judge choice and potential same-family preference, we additionally calibrate prompt of GPT-4o as an independent judge and use it to re-evaluate representative models. As shown in Table~\ref{tab:cross_judge}, the two judges produce highly consistent scores and overall performance patterns across model families. The LEVEL-1 ranking remains unchanged, while LEVEL-2 shows only minor rank permutations among the top-performing models. These results suggest that our main conclusions are robust to the choice of judge and are unlikely to be driven by same-family preference.

\subsection{RQ4: Can fine-grained evaluation metrics provide actionable guidance for improving model capabilities?}
To evaluate whether the proposed fine-grained intermediate reflection and decision metrics can provide effective supervision to improve agent performance, we use Qwen3-8B as the base model and conduct SFT and DPO~\citep{rafailov2023direct}, respectively. We first randomly construct a fixed test set. These test instances are excluded from all subsequent training data construction to avoid data leakage. The remaining tasks are used to construct the training data.

\textbf{Few-shot Prompting.} To examine how much capability can be elicited through prompting
alone, we additionally introduce a few-shot baseline by augmenting the original prompt with three high-quality demonstrations.

\textbf{Supervised Fine-Tuning.} For SFT, we collect trajectories generated by multiple models and score them using the SDE evaluation metrics. Trajectories with an average score greater than 10 (15 sub-dimensions in total, with a total score of 15) are selected as candidate high-quality trajectories and are further manually cleaned. This process results in 4,924 SFT training instances.

\textbf{ParaRecover-DPO.} Our SDE evaluation framework, as a point-wise scoring rubric, allows us to construct preferences based on score differences, thus making it compatible with preference-learning objectives such as DPO (and variants). Based on the SFT model, we further train with DPO. For DPO, we construct preference pairs from trajectories generated for the same task. The trajectory with a higher total score is treated as the preferred response, while the lower-scoring trajectory is treated as the rejected response. To ensure that the preference signal is reliable, we require all trajectories to satisfy a minimum threshold 3 and the score gap between the preferred and rejected trajectories to be greater than 5. After manual cleaning, we obtain 3,187 preference pairs for DPO training. 


As shown in Tab.~\ref{RL}, few-shot prompting
provides moderate improvements over the base model, achieving average scores of 44.47 and 42.24 on LEVEL-1 and LEVEL-2, respectively. This suggests that high-quality demonstrations can partially elicit better recovery behavior without parameter updates. Also, both SFT and ParaRecover-DPO improve the SDE scores, with ParaRecover-DPO achieving the best performance on both LEVEL-1 and LEVEL-2. 

Since the training data are constructed using the SDE rubric, however, improvements under the same rubric alone may not provide fully independent evidence of enhanced recovery capability.
We therefore further evaluate the models using judge-free, execution-based metrics, including tool-use efficiency, execution-round efficiency (E5), invalid DAG rate, and Pass@1. These metrics are computed directly from execution traces or environment outcomes and do not involve the LLM judge.

As shown in Tab.~\ref{tab:judge_free}, ParaRecover-DPO consistently improves all judge-free metrics across both difficulty levels. On LEVEL-1, tool-use efficiency increases from 20.43 to 26.20 and E5 from 36.50 to 40.85, while the invalid DAG rate decreases from 10.25 to 5.90 and Pass@1 rises from 91.84 to 94.30. Similar gains are also observed on LEVEL-2. These judge-independent results provide further evidence that SDE-guided preference learning improves actual execution and recovery behavior rather than merely fitting the judge's evaluation preferences.

\section{Conclusion}
We present ParaRecover, a novel process-level benchmark for evaluating error localization and recovery in parallel tool-use agents. Moving beyond previous benchmarks, our framework introduces a fine-grained error taxonomy, challenging error trajectories, and the SDE rubric to measure agent capabilities. Experimental results reveal that high task completion rates do not necessarily imply reliable recovery and replanning capabilities, especially under multi-turn error propagation. Furthermore, we demonstrate that SDE-based supervision can effectively improve agents’ reflective recovery behaviors.

\section{Limitation}
While ParaRecover offers a controlled and fine-grained framework for evaluating process-level error recovery in parallel tool-use agents, it has several limitations. First, our benchmark is built on a simulated execution environment rather than live real-world APIs. Although this design ensures reproducibility and controlled error injection, it may not fully capture the unpredictability, latency, and dynamic state changes inherent in real-world systems. Second, our SDE rubric employs an LLM-as-a-Judge paradigm to evaluate abstract reasoning dimensions. Despite achieving relatively high agreement with human annotations, this approach may still introduce bias in complex or ambiguous cases. Finally, ParaRecover models agent execution using DAG-structured workflows. While this formulation effectively captures dependency relationships in parallel multi-turn scenarios, it does not fully represent more dynamic real-world patterns such as loops, conditional branching, or asynchronous interactions. Future work may further improve ParaRecover through more realistic execution environments, hybrid human-AI evaluation protocols, and more general execution structures.

\section{Acknowledgements}
This work was supported in part by the National Natural Science Foundation
of China under Grant 62276044, and also in part by the 2025 Scientific Research Projects of the General Administration of Customs of China under Grant No. 2025HK184 and 2025HK209.

\bibliography{custom}

\appendix
\section{Related Work}

\subsection{LLM Agents}
LLM agents extend language models from text generators to goal-driven systems that can plan, act and interact with external environments~\citep{yao2022react, shinn2023reflexion}. An important direction is tool-augmented agents, where LLMs learn to invoke external search engines, calculators or domain-specific tools to overcome the limitations of parametric knowledge~\citep{schick2023toolformer, qin2024toolllm}. These design have made agents effective and flexible across domains. For example, research agents automate literature collection, scientific experiments, and paper writing~\citep{ren2025towards}; Code agents iteratively edit and debug in complex projects~\citep{yang2024swe, hong2024metagpt}; Web agents interact with browsers and operating systems to complete complex tasks~\citep{zhou2024webarena, he2024webvoyager}. These scenarios involve complex tool calls and long-range decision-making, therefore establishing high-quality benchmarks is crucial for expanding the capabilities of LLM agents.

\subsection{Agent Tool-use and Trajectory Evaluation.}

Existing tool-use benchmarks mainly evaluate tool selection, parameter filling, and task completion. API-Bank and Tool-Bench evaluate API retrieval, and execution over tool-use tasks~\citep{li2023api,qin2024toolllm}, while BFCL focuses on function-call correctness across real-world APIs, including serial and parallel function calls~\citep{patil2025berkeley}. More recent benchmarks such as GAIA, MCP-Bench and MCP-AgentBench extend evaluation to multi-step tool use, cross-tool coordination, protocol-mediated environments, and realistic agent workflows~\citep{mialon2024gaia,wang2025mcp,guo2026mcp}. However, these benchmarks are still mainly outcome-oriented or call-level oriented, making it difficult to diagnose where an agent fails within a long trajectory.

Recent trajectory-aware benchmarks attempt to provide more fine-grained evaluation. T-Eval decomposes tool utilization into instruction following, planning, reasoning, retrieval, understanding, and review~\citep{chen2024t}. TRAJECT-Bench evaluates whether tools are selected, parameterized, and ordered correctly along trajectories~\citep{he2025traject}. Reward-oriented benchmarks such as AgentRewardBench and Plan-RewardBench evaluate agent trajectory preferences~\citep{lu2025agentrewardbench, wang2026aligning}. Although these work moves beyond final-answer accuracy, they mainly focus on step correctness, tool-use correctness or trajectory-level preference judgment.

As shown in Table~\ref{comparison}, our work differs by targeting error-aware trajectory evaluation in parallel tool-calling scenarios. Instead of only asking whether a tool call is correct, whether the final answer is correct, or which trajectory is preferred, our benchmark evaluates whether an agent can inspect flawed intermediate trajectories, localize erroneous steps, and decide how to recover and replan. This setting complements existing tool-use and trajectory-evaluation benchmarks by focusing on the agent’s reflection and recovery decision ability under multi-turn parallel tool-use failures.

\begin{table}[t]
\centering
\small
\begin{tabular}{lcccc}
\toprule
\textbf{Model} & \multicolumn{2}{c}{\textbf{LEVEL-1}} & \multicolumn{2}{c}{\textbf{LEVEL-2}} \\
\cmidrule(lr){2-3} \cmidrule(lr){4-5}
& \textbf{E5} & \textbf{Pass@1} & \textbf{E5} & \textbf{Pass@1} \\
\midrule
GPT-4o                  & 42.19 & 92.21 & 40.26 & 91.50 \\
GPT-4o-mini             & 36.17 & 89.32 & 30.05 & 88.84 \\
Claude Opus 4.6         & 60.61 & 94.78 & 57.50 & 94.05 \\
Gemini 3.1 pro          & 52.81 & 93.68 & 49.32 & 93.22 \\
Qwen 3 8B               & 36.50 & 91.84 & 30.21 & 91.37 \\
Qwen 3 32B              & 39.94 & 92.08 & 34.10 & 91.86 \\
DeepSeek-V4-flash       & 57.42 & 94.13 & 55.68 & 93.17 \\
\bottomrule
\end{tabular}
\caption{E5 score and pass@1 performance of different models on LEVEL-1 and LEVEL-2}
\label{tab7}
\end{table}

\section{Analysis of the Discrepancy Between Pass@1 and ES Scores}
\label{Analysis of the Discrepancy Between Pass@1 and ES Scores}
Table~\ref{tab7} presents the disparity between E5 scores and Pass@1 performance for several representative models across different task levels. We observe that while Pass@1 metrics remain consistently high (over 88\%) across all models, E5 scores are generally low (the scores are shown after normalization to 0\~{}100), with significant performance gaps among models of different parameter sizes. Since the E5 metric quantifies the deviation of actual execution rounds from ground-truth rounds, 
the contrasting trends of these two metrics indicate that although models can complete our benchmark tasks within a limited number of rounds (see Appendix~\ref{times}), their execution processes are plagued by erroneous planning and redundant tool calls, leading to severe inefficiency and high costs.

\begin{figure*}[t]
  \includegraphics[width=\textwidth]{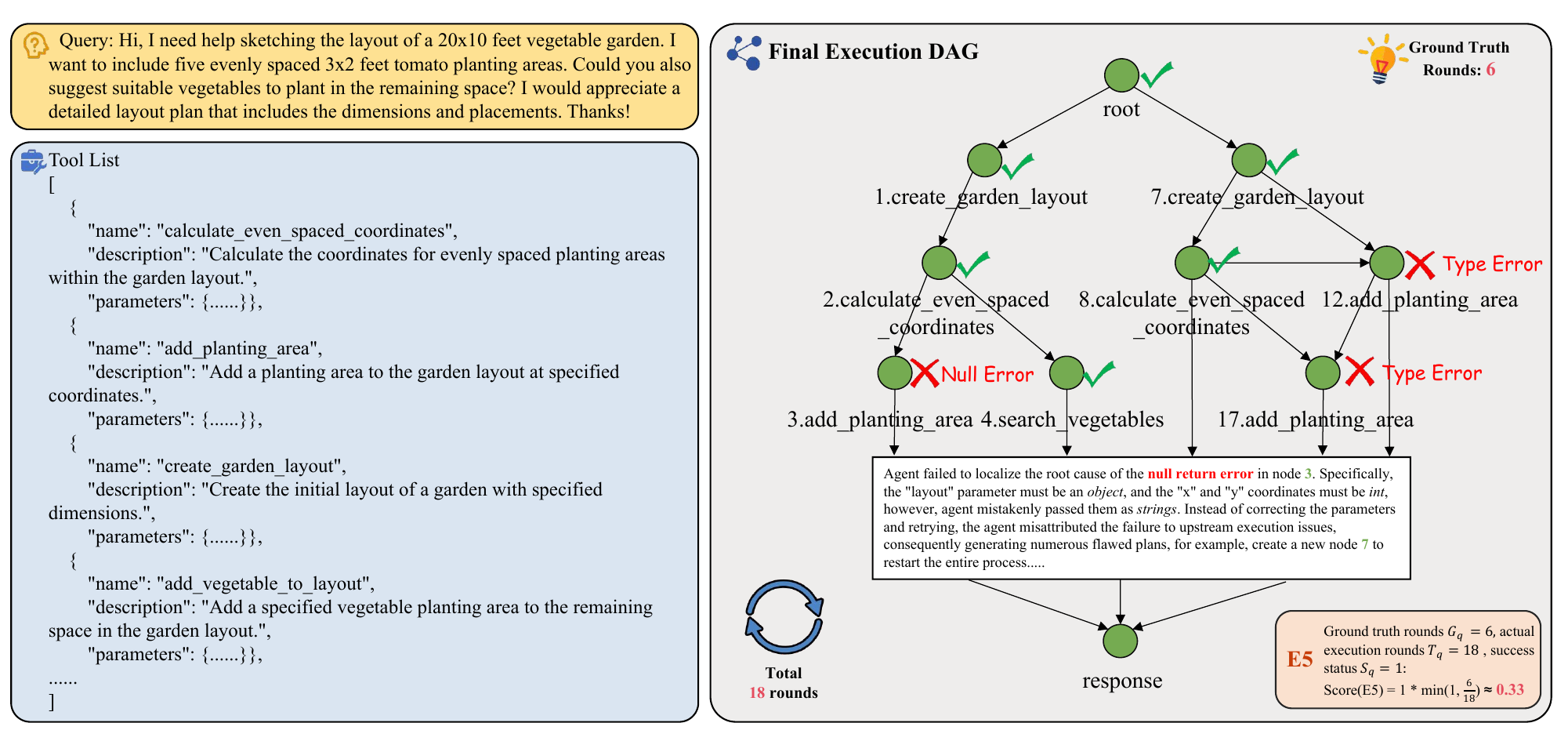}
  \caption{Case study}
  \label{fig5}
\end{figure*}

The case study in Fig.~\ref{fig5} illustrates this phenomenon. In this relatively simple task, Node 3 returns an empty response due to an incorrect parameter type. However, instead of performing valid error attribution and retrying the call, the agent misattributes the failure to an upstream task. Consequently, it creates Node 7 to retry the entire workflow, triggering a cascade of subsequent errors. The actual execution takes 18 rounds, compared to the ground-truth requirement of 6 rounds, resulting in an E5 score of 0.33. Such cases are prevalent, demonstrating that agents struggle with effective error attribution in multi-turn parallel scenarios, which frequently generates redundant steps. This reveals the limitations of relying solely on final task completion rates, underscoring the critical importance of evaluating agents' error localization and replanning capabilities when handling flawed trajectories.

\section{Error Description}
\label{Error Description}
Table~\ref{descriptions} shows the 14 error types and their corresponding descriptions.

\section{Error Type Distribution}
\label{Error Type Distribution}

\begin{figure}[htbp]
    \centering
    \includegraphics[width=\columnwidth, keepaspectratio]{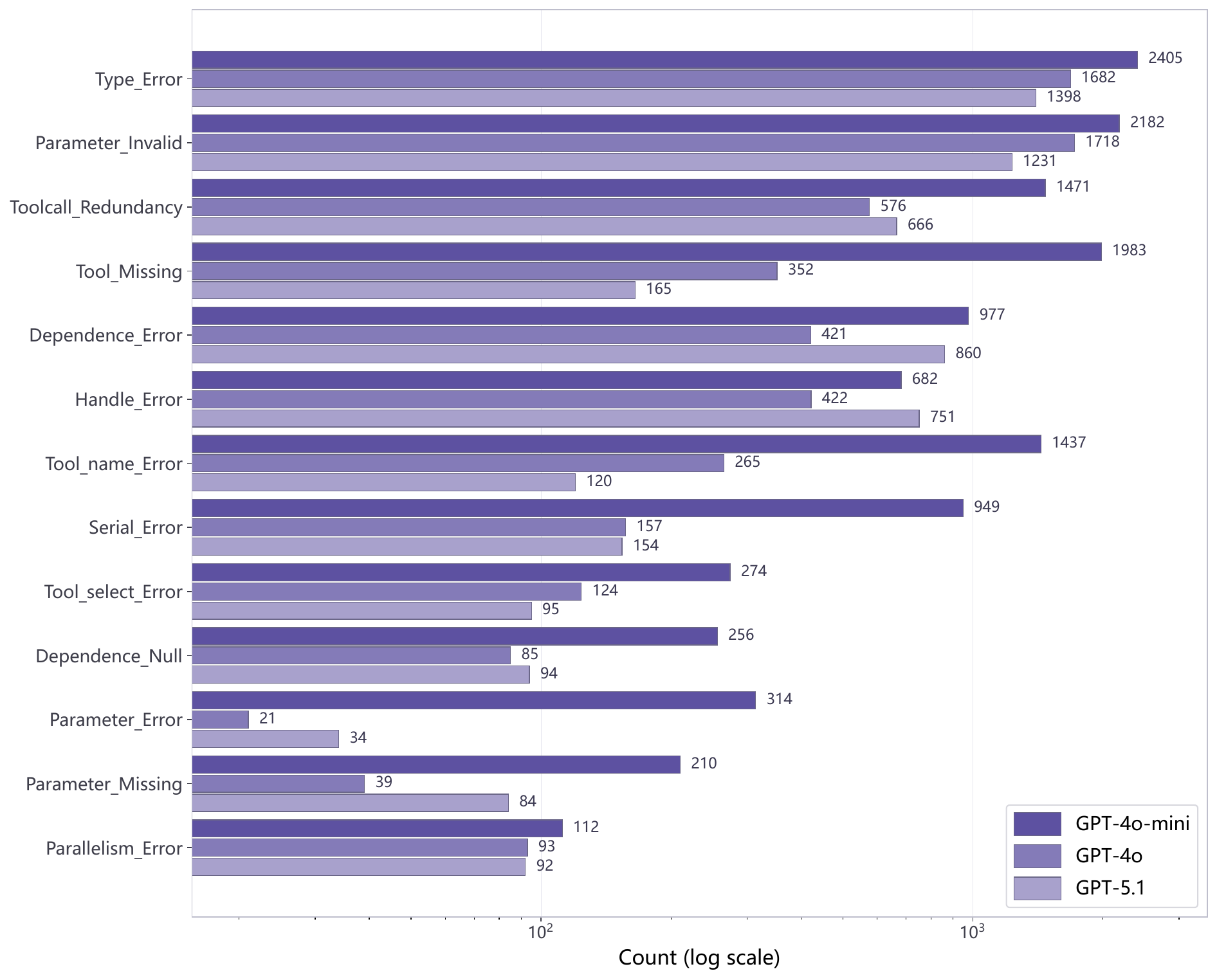}
    \caption{Comparative analysis of error type occurrences across GPT-4o-mini, GPT-4o, and GPT-5.1. The horizontal bars represent the frequency of each error category on a logarithmic scale.}
    \label{fig:error_comparison}
\end{figure}

As shown in Fig.~\ref{fig:error_comparison}, we analyze the error distributions of three different models during real-world execution. The most frequent errors encountered are incorrect parameter types, invalid parameter values, redundant tool calls, and missing critical tool calls. Notably, models with larger parameter counts and stronger reasoning capabilities consistently exhibit lower error rates compared to smaller models. This distribution also demonstrates that all error categories within our taxonomy occur in real-world agent execution scenarios.

\section{Related Prompts}
\label{prompt}
In this section, we present prompts used across various stages of our experiments. Specifically, these prompts comprise the DAG-based CoT agent execution planning prompt, the tool agent prompt used to simulate tool execution results in our controlled environment, and the LLM-as-a-Judge evaluation prompts across the three dimensions of the SDE rubric.
\begin{tcolorbox}[title = {Prompt for Agent},breakable,
    colback = white,
    colframe = blue!20,        
    coltitle = black,
    fonttitle = \bfseries,
    fontupper = \scriptsize,
    sharp corners,
    enhanced,
    drop shadow = blue!10!gray, 
    boxrule = 0.8pt, ] 
\ \ You are an expert at leveraging various tools to solve problems. You focus on observing the [Front-Wheel Plan] and [Front-Wheel Tool Call Results], and then make appropriate reflections and plans based on the [Original User Query] and [Available Tools List].
\\
\#\# Your task:
\\
1. Analyze the user's original query, combine it with the DAG list of the previous plan and the results of the previous tool calls, and consider what needs to be done to complete the task.
\\
2. Based on the tools that still need to be called, update the current complete task list to guide the entire subsequent process.
\\
\#\# Output requirements
\\
- First, output the <thought> section. Include reflections on the original user task and perform a requirement analysis: first, identify which subtasks have already been completed toward answering the user's question and which subtasks still need to be completed. Then, provide a detailed sequence of steps required to accomplish the original task, along with a sound justification for the rationale behind each step. In your reasoning, include a detailed analysis of the reasonableness of the previous round's planning, as well as a detailed analysis of the reasonableness of the previous round's tool invocation arrangements (including the tools used and their parameters).
\\
- Second, output a parseable JSON representing the task chain list of tool calls.
    1. Clearly specify the dependency relationships for each subtask. A dependency exists only if the return result of a preceding task is required to supplement or complete the parameters of a subsequent tool call. For any subtask that does not require preceding information, its dependency should be set to "root".
    2. For subtask nodes that have not yet been executed, you are free to modify, add, or delete them as needed to better invoke tools and complete the task. Note that subtask nodes which have already been executed must not be altered.
    3. The first subtask node has been determined as the "root" node, with the subtask set to:
\\

\{"id": "root",  \# Root node, representing the starting point of the task.
"desc": \{query\}, \# Original query content provided by the user.
"dep": [],  \# The root node does not need to depend on other nodes.
"status": True   \# The root node has already been executed by default.\}
    The last subtask node has been determined as the "response" node, with the subtask set to:
\{"id": "response",  \# The final node indicates the end point of the task.
"desc": "Analyze and summarize, and then respond.", \# Summarize and analyze the results of all subtasks, and then respond.
"dep": [...],  \# All nodes that are not dependent on by other subtasks.
"status": False    \# The last response node has not been executed by default.\}
\\
\#\#The output format is as follows (strictly output the reasoning section and the task list JSON. Do not include comments. Please output a compact JSON format, removing all unnecessary line breaks and indentation spaces, while preserving necessary spaces within string content):
\\
<thought>Output your thoughts on the user's current query; you need to re-analyze and rethink it. Conduct a requirements analysis, first analyzing what you have already accomplished to complete the user's original query, and what tools are still needed. Then, detail the steps required to answer the user's query, and reasonably explain the rationale for each step. Your thinking should include a detailed analysis of the rationality of the previous planning, as well as the rationality of the tool usage arrangement (including tools used and parameters). Note that your thinking should not exceed 500 words.</thought>

\{"tasks": [  \# A list of subtasks, where each element is a tool call subtask node, containing the task ID, the name of the called tool, the task description, and dependencies. All subtask nodes must be provided (regardless of whether they have been executed), including the root node and the response subtask node.
\{"id": string,  \# A unique identifier for a subtask, which is identified by a number "n", starting from "1".
"name": string,  \# The name of the tool called in the subtask.
"para": [string],  \# The parameters used in the tool call are in the format \{"parameter1 name":"parameter1 value","parameter2 name":"parameter2 value", ......\}.
"desc": string, \# Description of task nodes
"dep": [string],  \# The subtask dependencies, where each element in the list is the id of the dependent subtask.
"status": bool  \# This indicates whether the subtask has been executed. True means it has been executed, and False means it has not been executed.\}, ......]\}

\#\#Original User Query
\\
\{query\}
\\
\#\#Available Tools List
\\
\{tool\_list\}
\\
\#\#Front-Wheel Plan
\\
\{plan\_dag\}
\\
\#\#Front-Wheel Tool Call Results
\\
\{tool\_call\_dag\}

\end{tcolorbox}
\begin{tcolorbox}[title = {Prompt for Tool Agent},breakable,
    colback = white,
    colframe = blue!20,        
    coltitle = black,
    fonttitle = \bfseries,
    fontupper = \scriptsize,
    sharp corners,
    enhanced,
    drop shadow = blue!10!gray, 
    boxrule = 0.8pt, ] 
\ \ You are an expert skilled in simulating agent tool invocation. You need to generate simulated execution results based on the User Task and Available Tool List.
\\
\#\#Your Task
\\
Carefully identify the pending executable nodes in the previous planned DAG graph, namely nodes marked False with all dependency requirements satisfied.
Verify the validity of tool invocations in the DAG graph. Check whether tool names, parameter names and parameter types comply with standard specifications defined in the Available Tool List. Return reasonable error prompts immediately if any irregularities are detected.
If the invocation is valid, generate plausible returned results corresponding to the description and assigned tool of each task node.
All simulated results shall be logical and consistent with real-world scenarios.
Deliver accurate calculation outputs for computational tools.
Return reasonable summarized search contents for retrieval tools. Realistic fabricated values associated with the given task are acceptable.
\\
\#\#DAG Graph Description
\\
The previous plan is presented as a DAG graph starting from the root node and ending at the response node.
DAG structure format:
\\
\{"tasks": [\{"id": string, "name": string, "para": \{\}, "desc": string, "dep": [string], "status": bool\}, ...]\}
Nodes ready for execution are unexecuted nodes with False status, whose all dependent predecessor nodes have been fully completed with True status.
Output Format
First write your reasoning process within <thought></thought>.Then output tool responses in standard JSON array format:
'''json
[
    \{
        "name": string,  \# Consistent with the name field in DAG graph
        "arguments": \{"Parameter1": "Value1"\},  \# Tool calling parameters
        "results": string  \# Simulated tool execution feedback
    \},
    ...
]
\\
\#\#User Task
\\
\{query\}
\\
\#\#Available Tool List
\\
\{tool\_list\}
\\
\#\#Previous Planned DAG Graph
\\
\{replan\_dag\}

\end{tcolorbox}
\begin{tcolorbox}[title = {Prompt for Structural Integrity Evaluation},breakable,
    colback = white,
    colframe = blue!20,        
    coltitle = black,
    fonttitle = \bfseries,
    fontupper = \scriptsize,
    sharp corners,
    enhanced,
    drop shadow = blue!10!gray, 
    boxrule = 0.8pt, ] 
\#\#Role
\\
You are a highly strict, conservative, and fine-grained evaluation expert for replanner decision-making.
\\
\#\#General Requirements
\\
You must evaluate whether the given replanner’s thought and replan DAG are truly high-quality.You must default to strict scoring rather than moderate or high scores by default.
\\
\#\#Evaluation Dimension: Structural Integrity 
\\
Strictly evaluate the following 4 sub-items judged by LLM from the perspective of plan structure and executability consistency.An additional script-based judgment item will be automatically supplemented in the program:
\\
\#\#S1 Thought-DAG Consistency: Whether the actions, repairs, and progress claimed in the thought are truly reflected in the replan DAG.
\\
1: Key repair actions, downstream processing, and DAG nodes in the thought correspond one-to-one.
\\
0.5: Partially consistent, but with omissions, misalignments, or claims only in the thought without implementation in the DAG.
\\
0: Clearly inconsistent.
\\
S2 Topological Dependency \& Overall Executability: Whether node dependencies, execution order, and upstream/downstream relationships are correct, and whether the overall structure forms a real executable DAG.
1: Correct and reasonable dependencies with strong overall executability.
0.5: Minor dependency issues or execution risks exist, but the main chain is generally understandable.
0: Obvious dependency errors or overall difficulty in execution.
\\
S3 Tool Call Legitimacy: Whether tool names, parameter names, and parameter value formats strictly comply with the constraints in the tool list.
1: All tool names, parameter names, and parameter value formats are legal and accurate.
0.5: Minor non-standard formats or potential risks exist, but not completely unusable.
0: Invalid tools, wrong parameter names, missing key parameters, or obviously illegal values.
\\
S4 State Progression Correctness: Whether the statuses of executed nodes, unexecuted nodes, retained nodes after failure, and response nodes are reasonable.
1: State annotations are consistent with the actual execution status.
0.5: Minor non-standard state annotations.
0: Obvious state errors that will mislead the executor.
\\
\#\#Output Format
\\
Output ONLY a JSON object, NO extra content:
'''json
\\
\{
  "S1\_score": "1 / 0.5 / 0",
  "S1\_reason": "Reason for this score",
  "S2\_score": "1 / 0.5 / 0",
  "S2\_reason": "Reason for this score",
  "S3\_score": "1 / 0.5 / 0",
  "S3\_reason": "Reason for this score",
  "S4\_score": "1 / 0.5 / 0",
  "S4\_reason": "Reason for this score",
\}
\\
'''
\\
\#\#Input
\\
\#\#Original User Task
\\
\{query\}
\\
\#\#Available Tool List
\\
\{tool\_list\}
\\
\#\#Previous Plan
\\
\{prev\_plan\}
\#\#Execution Results \& Tool Responses of Previous Plan
\\
\{tool\_result\}
\\
\#\#Output of ReplannerThought
\\
\{thought\}
\\
\#\#Updated Plan
\\
\{new\_plan\}

\end{tcolorbox}

\begin{tcolorbox}[title = {Prompt for Diagnostic Reasoning Evaluation},breakable,
    colback = white,
    colframe = blue!20,        
    coltitle = black,
    fonttitle = \bfseries,
    fontupper = \scriptsize,
    sharp corners,
    enhanced,
    drop shadow = blue!10!gray, 
    boxrule = 0.8pt, ] 
\#\#Role
\\
You are a highly strict, conservative, and fine-grained evaluation expert for replanner decision-making.
\\
\#\#General Requirements
\\
You must evaluate whether the given replanner’s thought and replan DAG are truly high-quality.You must default to strict scoring rather than moderate or high scores by default.
\\
\#\#Evaluation Dimension: Diagnostic Reasoning
\\
Strictly assess the following five sub-indicators from the perspective of diagnostic quality.
\\
D1 Plan Evidence AnchoringWhether the thought explicitly references and accurately understands specific nodes, dependencies, statuses or steps in the previous plan.
1: Clearly cites verifiable details such as node IDs, upstream and downstream nodes and status values, and conducts problem analysis based on valid evidence.
0.5: Contains partial references but fails to specify verifiable node IDs, dependencies or statuses, or lacks precise supporting evidence.
0: Barely relies on plan evidence and only provides general descriptions.
\\
D2 Tool Evidence AnchoringWhether the thought explicitly references and correctly interprets specific error messages, return values, parameter names or tool behaviors from tool results.
1: Precisely specifies exact tool names, parameter names, error fields or return values, and explains their implications.
0.5: Detects existing tool abnormalities but fails to reference specific parameters or error details, or presents incomplete evidence.
0: Makes no reference to concrete tool outputs or contains obvious misinterpretation.
\\
D3 Root Cause Localization AccuracyWhether the actual fundamental cause is identified instead of merely describing superficial symptoms.
1: Points out failure manifestations and accurately locates essential causes including invalid parameters, faulty dependencies and improper state transition.
0.5: Roughly pinpoints problematic sections but stays at superficial descriptions such as failed execution and parameter adjustment demand.
0: Delivers incorrect root cause judgment or fails to identify underlying reasons.
\\
D4 Impact Scope RecognitionWhether subsequent nodes, dependency chains, state transitions and response generation affected by errors are recognized.
1: Clearly names at least one affected downstream node, relevant dependency chain or abnormal response progress.
0.5: Acknowledges potential impacts but fails to specify affected nodes or covers incomplete influence scope.
0: Fails to identify any affected scope.
\\
D5 Evidence Sufficiency and RestraintWhether reasoning is sufficiently supported by facts without fabricating non-existent information.
1: Draws conclusions based on at least two independent verifiable evidence sources such as plan and tool results; actively detects remaining flaws and risks; distinguishes authentic root causes from irrelevant misleading explanations.
0.5: Partially meets requirements but shows noticeable deficiencies in evidence adequacy, false cause elimination or residual risk detection.
0: Lacks solid supporting grounds, contains obvious speculation and fabrication, or lacks rigorous verification awareness.
\\
\#\#Output Format
\\
Output only a single JSON object with no extra content.
'''json
\\
\{
  "D1\_score": "1 / 0.5 / 0",
  "D1\_reason": "State the reason for the score",
  "D2\_score": "1 / 0.5 / 0",
  "D2\_reason": "State the reason for the score",
  "D3\_score": "1 / 0.5 / 0",
  "D3\_reason": "State the reason for the score",
  "D4\_score": "1 / 0.5 / 0",
  "D4\_reason": "State the reason for the score",
  "D5\_score": "1 / 0.5 / 0",
  "D5\_reason": "State the reason for the score",
\}
\\
'''
\\
\#\#Input
\\
\#\#Original User Task
\\
\{query\}
\\
\#\#Available Tool List
\\
\{tool\_list\}
\\
\#\#Previous Plan
\\
\{prev\_plan\}
\#\#Execution Results \& Tool Responses of Previous Plan
\\
\{tool\_result\}
\\
\#\#Output of ReplannerThought
\\
\{thought\}
\\
\#\#Updated Plan
\\
\{new\_plan\}

\end{tcolorbox}

\begin{tcolorbox}[title = {Prompt for Diagnostic Evolutionary Strategy},breakable,
    colback = white,
    colframe = blue!20,        
    coltitle = black,
    fonttitle = \bfseries,
    fontupper = \scriptsize,
    sharp corners,
    enhanced,
    drop shadow = blue!10!gray, 
    boxrule = 0.8pt, ] 
\#\#Role
\\
You are a highly strict, conservative, and fine-grained evaluation expert for replanner decision-making.
\\
\#\#General Requirements
\\
You must evaluate whether the given replanner’s thought and replan DAG are truly high-quality.You must default to strict scoring rather than moderate or high scores by default.
\\
\#\#Evaluation Dimension: Evolutionary Strategy
Strictly assess the following five sub-items in terms of the quality of revision and improvement strategies.
\\
E1 Minimal ModificationWhether the replan only revises necessary parts without irrelevant rewriting.
1: Only essential nodes and dependencies are adjusted, and changes directly correspond to specific errors.
0.5: Revisions follow a general correct direction yet remain superficial, with redundant or excessive adjustments lacking sufficient justification.
0: Uncontrolled modifications made or key issues left unaddressed.
\\
E2 Closed-loop RevisionWhether the revision fixes current errors and complements relevant downstream nodes, parameters, statuses and response chains.
1: A complete closed loop is formed with explicit handling of affected downstream nodes, parameters and statuses.
0.5: Partial errors get fixed, yet gaps remain in downstream links, status transition and response chains.
0: No executable closed loop established.
\\
E3 Goal ConsistencyWhether the updated plan still faithfully fulfills the original user task without deviation, incomplete delivery or missing core objectives.
1: Original goals fully retained with complete pathways reserved for final response generation.
0.5: Basic goals maintained but partial coverage missing or minor deviation exists, and complete final response loop cannot be guaranteed.
0: Noticeable deviation or omission of core objectives.
\\
E4 Strategy RationalityWhether added, removed or reordered steps and dependencies possess clear practical necessity.
1: All strategic adjustments are fully justified and necessary.
0.5: Most adjustments are reasonable while a few changes lack adequate necessity.
0: Critical adjustments serve no practical purpose and appear as mechanical rewriting.
\\
\\
\#\#Output Format
\\
Output only a single JSON object with no extra content.
\\
'''json
\\
\{
  "E1\_score": "1 / 0.5 / 0",
  "E1\_reason": "State the reason for the score",
  "E2\_score": "1 / 0.5 / 0",
  "E2\_reason": "State the reason for the score",
  "E3\_score": "1 / 0.5 / 0",
  "E3\_reason": "State the reason for the score",
  "E4\_score": "1 / 0.5 / 0",
  "E4\_reason": "State the reason for the score",
\}
\\
\#\#Input
\\
\#\#Original User Task
\\
\{query\}
\\
\#\#Available Tool List
\\
\{tool\_list\}
\\
\#\#Previous Plan
\\
\{prev\_plan\}
\#\#Execution Results \& Tool Responses of Previous Plan
\\
\{tool\_result\}
\\
\#\#Output of ReplannerThought
\\
\{thought\}
\\
\#\#Updated Plan
\\
\{new\_plan\}
\end{tcolorbox}

\section{Data Distribution}
\label{Data Distribution}
\begin{figure}[t]
  \centering    \includegraphics[width=0.85\columnwidth,height=0.25\textheight]{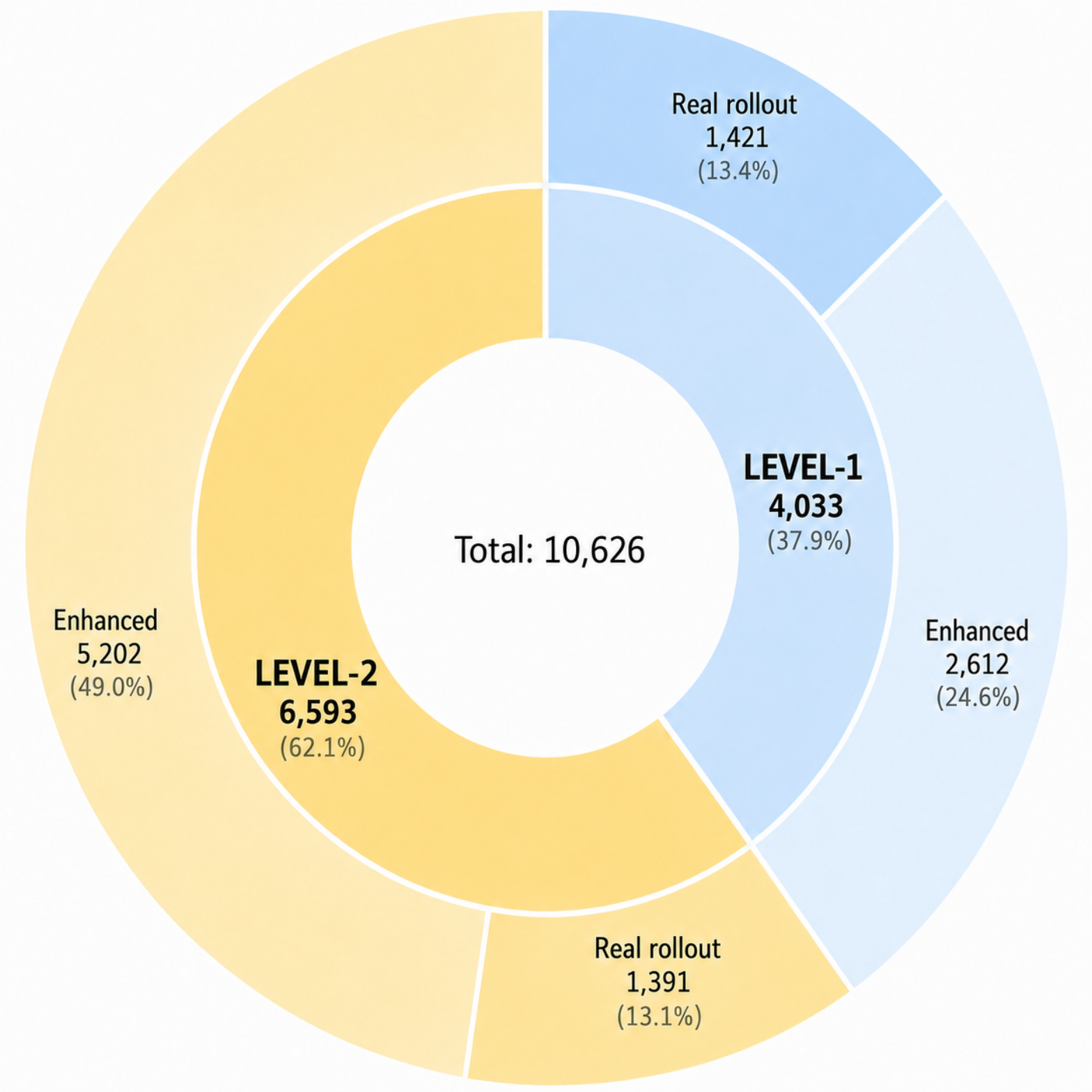}  
  \caption{Data distribution}
  \label{cmark1}
\end{figure}

Fig.~\ref{cmark1} illustrates the data distribution of our constructed benchmark. The benchmark comprises a total of 10,626 instances categorized into two levels: LEVEL-1 and LEVEL-2. Specifically, LEVEL-1 contains 4,033 instances, including 1,421 error instances collected from real-world rollouts and 2,612 instances generated via synthetic error augmentation on correct trajectories. LEVEL-2 consists of 6,593 instances, comprising 1,391 error instances from real-world rollouts and 5,202 instances derived through error augmentation on correct trajectories. Table~\ref{LEVEL1} and Table~\ref{LEVEL2} present the data distribution across different error types within the LEVEL-1 and LEVEL-2 tasks of ParaRecover. Specifically, LEVEL-1 tasks contain only single-round errors, whereas LEVEL-2 tasks encompass multi-turn error types.

\begin{table}[htbp]
\centering
\small
\begin{tabularx}{\linewidth}{l S[table-format=4.0] S[table-format=5.2]}
\toprule
\textbf{Error Type} & \textbf{Frequency} & \textbf{Ratio(\%)} \\
\midrule
Parameter\_Missing      & 333 &  8.26 \\
Parameter\_Invalid      & 341 &  8.46 \\
Parameter\_Error        & 276 &  6.84 \\
Type\_Error             & 322 &  7.98 \\
Tool\_name\_Error       & 258 &  6.40 \\
Tool\_select\_Error     & 266 &  6.60 \\
Toolcall\_Missing       & 307 &  7.61 \\
Toolcall\_Redundancy    & 296 &  7.34 \\
Dependence\_Error       & 278 &  6.89 \\
Dependence\_Null        & 252 &  6.25 \\
Serial\_Error           & 407 & 10.09 \\
Parallelism\_Error      & 268 &  6.65 \\
Null\_Error             & 210 &  5.20 \\
Timeout\_Error          & 219 &  5.43 \\
\midrule
\multicolumn{1}{c}{Total} & 4033 & 100.00 \\
\bottomrule
\end{tabularx}
\caption{Error type statistics in LEVEL-1}
\label{LEVEL1}
\end{table}

\begin{table*}
\centering
\footnotesize   
\begin{tabular}{l *{6}{r}}
\toprule
\textbf{Error Type} & 
\textbf{Round1} & \textbf{Ratio(\%)} & 
\textbf{Round2} & \textbf{Ratio(\%)} & 
\textbf{Round3} & \textbf{Ratio(\%)} \\
\midrule
Parameter\_Missing      & 429 &  6.51 & 443 &  6.72 & 198 &  7.20 \\
Parameter\_Invalid      & 528 &  8.01 & 544 &  8.25 & 204 &  7.42 \\
Parameter\_Error        & 428 &  6.49 & 416 &  6.31 & 187 &  6.80 \\
Type\_Error             & 598 &  9.07 & 553 &  8.39 & 197 &  7.16 \\
Tool\_name\_Error       & 554 &  8.40 & 485 &  7.36 & 181 &  6.58 \\
Tool\_select\_Error     & 435 &  6.60 & 370 &  5.61 & 177 &  6.44 \\
Toolcall\_Missing       & 558 &  8.46 & 554 &  8.40 & 203 &  7.38 \\
Toolcall\_Redundancy    & 540 &  8.19 & 530 &  8.04 & 206 &  7.49 \\
Dependence\_Error       & 449 &  6.81 & 483 &  7.33 & 208 &  7.56 \\
Dependence\_Null        & 395 &  5.99 & 417 &  6.32 & 203 &  7.38 \\
Serial\_Error           & 506 &  7.67 & 494 &  7.49 & 191 &  6.95 \\
Parallelism\_Error      & 353 &  5.35 & 397 &  6.02 & 214 &  7.78 \\
Null\_Error             & 394 &  5.98 & 456 &  6.92 & 189 &  6.87 \\
Timeout\_Error          & 426 &  6.46 & 451 &  6.84 & 191 &  6.95 \\
\midrule
\textbf{Total}          & 6593 & 100.00 & 6593 & 100.00 & 2750 & 100.00 \\
\bottomrule
\end{tabular}
\caption{Error type statistics in LEVEL-2}
\label{LEVEL2}
\end{table*}

\section{Training Configuration}

\label{train}
All experiments were conducted on a single server equipped with 8 NVIDIA RTX 3090 GPUs. We adopted Low-Rank Adaptation (LoRA)~\citep{hu2022lora} for both the SFT and DPO stages. The model used is Qwen3-8B. The LoRA hyperparameters were uniformly set to a rank of $r=32$, a scaling factor of $\alpha=64$, and a dropout rate of $0.05$. For the SFT stage, we trained the base model for 3 epochs with a maximum sequence length of 4096 tokens to accommodate multi-turn agent execution trajectories. We employed a cosine learning rate scheduler with a peak learning rate of $1 \times 10^{-4}$ and a 3\% warmup ratio. The global batch size was set to 64, achieved through a per-device batch size of 2 and 4 gradient accumulation steps. For the DPO stage, we initialized the policy and reference models with the resulting SFT weights. To prevent overfitting on the preference data, DPO training was limited to 1 epoch. The learning rate was reduced to $5 \times 10^{-6}$ with a cosine scheduler and a 10\% warmup ratio. The KL penalty coefficient was set to $\beta=0.1$, and the global batch size was maintained at 64.

\begin{table*}[t]
  \small
  \centering
  \begin{tabular}{p{0.25\linewidth} p{0.68\linewidth}}
    \toprule
    \textbf{Error Type} & \textbf{Description} \\
    \midrule
    Dependence\_Error & Dependency Relation Error: In the unfinished tasks of the updated DAG generated by the agent, some dependencies between nodes are incorrectly specified. For example, node 3 should depend on node 2, but is mistakenly represented as depending on node 1. \\
    \midrule
    Dependence\_Null & Missing Dependency Error: In the unfinished tasks of the updated DAG generated by the agent, some nodes have missing dependencies. \\
    \midrule
    Parallelism\_Error & Incorrect Parallelization: In the unfinished tasks of the updated DAG generated by the agent, some nodes are incorrectly arranged in parallel. Specifically, two nodes that should be executed sequentially are treated as parallel. For example, node 2 requires the execution result of node 1 as its input parameter, but the agent schedules them to run in parallel. \\
    \midrule
    Serial\_Error & Incorrect Serialization: In the unfinished tasks of the updated DAG generated by the agent, some nodes are incorrectly arranged sequentially. Specifically, two nodes have no dependency on each other and can be executed in parallel, but they are mistakenly represented as having a sequential dependency in the updated DAG. \\
    \midrule
    Parameter\_Error & Parameter Name Error: In the unfinished tasks of the updated DAG generated by the agent, some tool-call nodes contain invalid parameters that are not defined in the current tool’s parameter list. For example, the supported parameter in the parameter list is $begin\_date$, but the tool-call node incorrectly provides $begin-date$. \\
    \midrule
    Parameter\_Invalid & Invalid Parameter Value: In the unfinished tasks of the updated DAG generated by the agent, some tool-call nodes assign parameter values with the correct type but invalid content. For example, the tool requires a $date$ parameter, but the call provides an invalid value such as "2026/02/32". \\
    \midrule
    Parameter\_Missing & Missing Required Parameter: Each tool is associated with a parameter list and a set of required parameters. In the unfinished tasks of the updated DAG generated by the agent, some tool-call nodes fail to provide required parameters. \\
    \midrule
    Type\_Error & Parameter Type Error: Each parameter in a tool’s parameter list has a predefined value type, such as "int" or "string". In the unfinished tasks of the updated DAG generated by the agent, some tool-call nodes assign values with incorrect types. For example, a parameter requires the integer value $16$, but the node provides the string value $``16"$. \\
    \midrule
    Null\_Error & Empty Tool Response Error: During tool execution, the tool returns an empty result for some reason. \\
    \midrule
    Timeout\_Error & Timeout Error: During tool execution, the tool call fails with a timeout error due to issues such as network instability. \\
    \midrule
    Tool\_name\_Error & Invalid Tool Name Error: In the unfinished tasks of the updated DAG generated by the agent, some tool-call nodes invoke an invalid tool name that is not included in the current tool list. \\
    \midrule
    Tool\_select\_Error & Incorrect Tool Selection: Some tools in the tool list have similar or easily confused functions. In the unfinished tasks of the updated DAG generated by the agent, some tool-call nodes select an inappropriate tool for the intended operation. \\
    \midrule
    Toolcall\_Missing & Missing Required Tool Call: In the updated plan DAG generated by the agent, some tool calls required to complete the task are missing. \\
    \midrule
    Toolcall\_Redundancy & Tool call redundancy: In the unfinished tasks of the updated DAG generated by the agent, there are some unnecessary tool calls. \\
    \bottomrule
  \end{tabular}
  \caption{Error Types and Descriptions}
  \label{descriptions}
\end{table*}

\section{Maximum Number of Execution Rounds}
\label{times}
In our experiments, an agent is considered to have successfully completed a task if it reaches the response node within the maximum number of execution rounds, $r$. We set $r = 40$ because preliminary experiments show that trajectories exceeding 40 rounds typically indicate either an inherent bias in the agent's task comprehension or an infinite loop. Consequently, such tasks are highly unlikely to be completed even if additional rounds are allowed.

\end{document}